%% file: 000-main.tex
\documentclass{article} % For LaTeX2e
\usepackage{iclr2024_conference,times}

\usepackage{hyperref}
\usepackage{url}

\usepackage{booktabs}       % professional-quality tables
\usepackage{amssymb}
\usepackage{amsfonts}       % blackboard math symbols
\usepackage{nicefrac}       % compact symbols for 1/2, etc.
\usepackage{microtype}      % microtypography
\usepackage{xcolor}         % colors
\usepackage{xspace}
\usepackage{subcaption}
\usepackage{graphicx}
\usepackage{wrapfig}
\usepackage{algorithm}
\usepackage{algorithmic}
\usepackage{enumitem}
\usepackage{dsfont}
\usepackage{bbm}
\usepackage{listings}
\usepackage{array}
\usepackage{inconsolata}

\input{sections/notations}
\input{sections/math_commands}

\title{Progressively Efficient Learning}

\author{\thanks{The first two authors contribute equally. Correspondence email:  \texttt{kxnguyen@berkeley.edu}.}$^*$Ruijie Zheng$^{\diamondsuit}$,
\textbf{$^*$}Khanh Nguyen$^{\clubsuit}$ ,
Hal Daum\'e III$^{\diamondsuit \heartsuit}$, Furong Huang$^{\diamondsuit}$,
Karthik Narasimhan$^{\spadesuit}$ \\
$^{\diamondsuit}$ University of Maryland, College Park \ \ \ \ $^{\heartsuit}$ Microsoft Research \\
$^{\clubsuit}$ University of California, Berkeley \ \ \ \ 
$^{\spadesuit}$ Princeton University 
}

\iclrfinalcopy % Uncomment for camera-ready version, but NOT for submission.
\begin{document}

\maketitle

\begin{abstract}
\input{sections/abstract}

\end{abstract}

\input{sections/intro}

\input{sections/motivation}

\input{sections/overview}

\input{sections/implementation}

\input{sections/exp_setup}
\input{sections/results}
\input{sections/related}
\input{sections/conclusion}

\bibliographystyle{iclr2024_conference}
\bibliography{iclr2024_conference,paper}

\newpage
\appendix
\input{appendix/a1_subtask_graph}
\input{appendix/a2_ablation_cost}
\input{appendix/a5_implementation}

\input{appendix/a6_teachers}

\end{document}

%% file: sections/notations.tex
\definecolor{mygreen}{HTML}{00890b}
\definecolor{mygray}{HTML}{9a9a9a}
\definecolor{mygold}{HTML}{705107}

\renewcommand{\sectionautorefname}{\S\kern-0.2em}
\renewcommand{\subsectionautorefname}{\S\kern-0.2em}

\newcommand{\ouralgo}{CEIL\xspace}

\newcommand{\doact}{\texttt{[do]}\xspace}
\newcommand{\doneact}{\texttt{[done]}\xspace}

\newcommand{\agentpol}{\pi_{\theta}}

%% file: sections/math_commands.tex
\usepackage{amsmath,amsfonts,bm}

\def\eqref#1{equation~\ref{#1}}
\def\1{\bm{1}}
\DeclareMathAlphabet{\mathsfit}{\encodingdefault}{\sfdefault}{m}{sl}
\SetMathAlphabet{\mathsfit}{bold}{\encodingdefault}{\sfdefault}{bx}{n}
\def\gA{{\mathcal{A}}}

\def\gG{{\mathcal{G}}}

\def\gI{{\mathcal{I}}}

\def\gS{{\mathcal{S}}}

\newcommand{\E}{\mathbb{E}}

\DeclareMathOperator*{\argmax}{arg\,max}

%% file: sections/abstract.tex
Assistant AI agents should be capable of rapidly acquiring novel skills and adapting to new user preferences.
Traditional frameworks like imitation learning and reinforcement learning do not facilitate this capability because they support only low-level, inefficient forms of communication.
In contrast, humans communicate with \textit{progressive efficiency} by defining and sharing abstract intentions. 
Reproducing similar capability in AI agents, we develop a novel learning framework named \textit{\textbf{C}ommunication-\textbf{E}fficient \textbf{I}nteractive \textbf{L}earning} (\ouralgo). 
By equipping a learning agent with an abstract, dynamic language and an intrinsic motivation to learn with minimal communication effort, \ouralgo leads to emergence of a human-like pattern where the learner and the teacher communicate progressively efficiently by exchanging increasingly more abstract intentions. 
\ouralgo demonstrates impressive performance and communication efficiency on a 2D MineCraft domain featuring long-horizon decision-making tasks. 
Agents trained with \ouralgo quickly master new tasks, outperforming non-hierarchical and hierarchical imitation learning by up to 50\% and 20\% in absolute success rate, respectively, given the same number of interactions with the teacher. 
Especially, the framework performs robustly with teachers modeled after human pragmatic communication behavior.
\looseness=-1

%% file: sections/intro.tex
\section{Introduction}

Imagine Alice, a programming expert, teaching Bob, a novice, how to write computer programs.
Initially, because they share little common knowledge in this domain, Alice has to demonstrate step by step how a program is written. 
This strategy quickly enables Bob to write simple programs, but it is inadequate for teaching him to compose sophisticated programs consisting of thousands of lines of code. 
Hence, after teaching through demonstrations for a while, Alice switches to a more efficient strategy.
She grows a shared vocabulary with Bob and gradually adds to it increasingly more abstract terms, which help them express complex intentions succinctly.  
For example, after explaining the concepts of ``\textit{for-loop}'' and ``\textit{a function that checks whether an integer is a prime}'', Alice can teach Bob to count the number of two-digit primes by giving an instruction like ``\textit{write a {\normalfont for loop} from 1 to 99, and call the {\normalfont prime-checking function} in each iteration}'' instead of having to dictate a full program to him.
In general, as Alice and Bob communicate more frequently and want to exchange increasingly complex ideas, they leverage existing shared terms to ground new, more abstract terms to reduce their communication effort. 
We refer to this phenomenon as \textit{progressively efficient communication}.

\begin{figure}[t!]
    \centering
    \includegraphics[width=\textwidth]{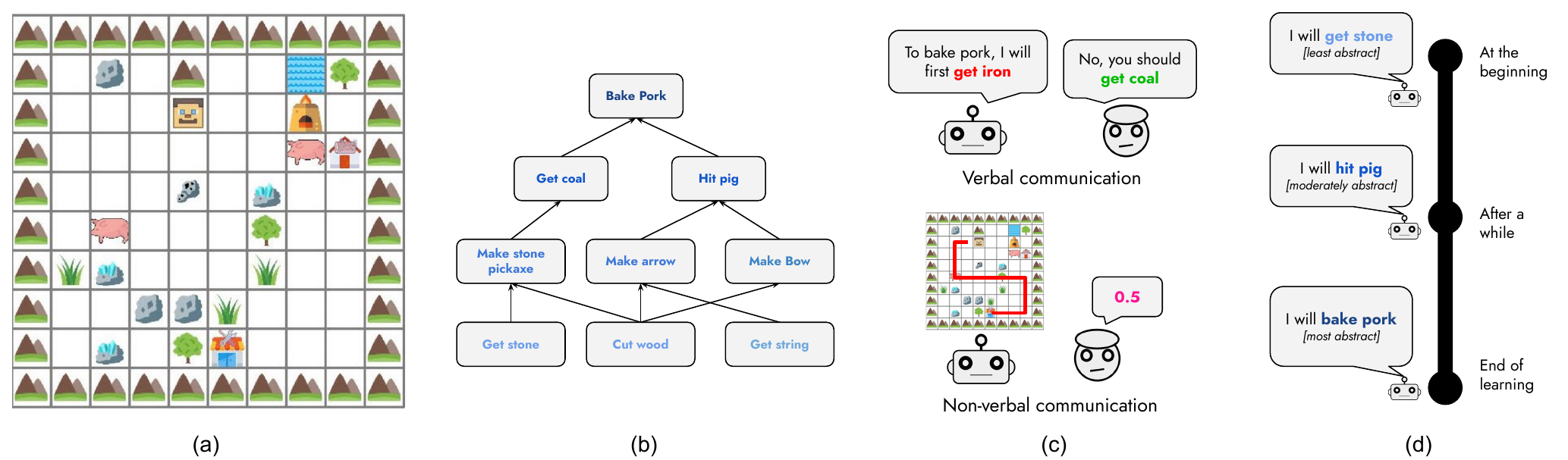}
         \label{sfig:demo1}
\caption{An illustration of \ouralgo on a ``bake pork'' task in a 2D MineCraft environment (a). The environment features compositional tasks which represent intentions at various levels of abstraction (b). \ouralgo enables the learner to quickly learn complex tasks by capitalizing on its mastery of simpler tasks. When interacting with the teacher (c), the learner can choose to verbally communicate an intention (a reference to an action sequence) and receive correction, or execute an intention by taking actions in the environment and obtain evaluation.
It can alternate between these two modes of communication within a learning episode.
Aiming to reduce communication effort, the learner conveys increasingly more abstract intentions over time (d). Learning efficiency enhances as communication becomes more abstract. \looseness=-1}
\label{fig:demo}
\end{figure}

In order to excel as personal assistants of humans, AI agents should be capable of progressively efficient communication.
These agents should handle increasingly complex user requests without demanding extensive user effort to adapt their behavior.
In this paper, we demonstrate that incorporating elements of human communication allows for the construction of such agents. 
We first identify two elements that are prerequisite for progressively efficient communication but are missing in traditional frameworks like imitation learning (IL) and reinforcement learning (RL): (i) a dynamic, referential communication medium (the means) and (ii) a desire to minimize collaborative effort (the motivation).  
We develop \textit{\textbf{C}ommunication-\textbf{E}fficient \textbf{I}nteractive \textbf{L}earning}  (\ouralgo), a learning framework that equips the learning agent with human-like means and motivation for progressively efficient communication.
As illustrated in \autoref{fig:demo}, \ouralgo transcends IL and RL by allowing the teacher and the learner to exchange abstract intentions rather than low-level signals like numerical rewards or primitive actions.
Furthermore, it injects into the learner an intrinsic motivation to minimize long-term communication effort, encouraging the learner to understand and use abstract terms to express intentions concisely. 
While incorporating one of these elements of human communication has been previously explored \citep{kulkarni2016hierarchical,le2018hierarchical,ren2021survey,brantley2020active,zhang2016query}, our work is the first to integrate \textit{both} of them in a single framework with the goal of mimicking the progressive efficiency of human communication.

We present a variant of our framework with IL-like instructive feedback and RL-like evaluative feedback, and propose a practical learning algorithm by extending Q-learning.
We evaluate the effectiveness of our algorithm on a challenging 2D MineCraft domain where each task is composed of various subtasks and requires a long sequence of actions to complete. 
Results indicate that our algorithm \ouralgo learns significantly faster and achieves higher asymptotic performance than various RL and IL baselines. 
The algorithm achieves performance gain consistently across various models of the teacher we construct, including those mimicking human pragmatic behavior.
The temporal change in the distribution of utterances indicates that progressively efficient communication indeed emerges, as the learner conveys increasingly more abstract intentions to the teacher.
Our work illustrates that integrating human communication traits holds great promise in the advancement of more efficient and human-compatible learning frameworks.

%% file: sections/motivation.tex
\section{Preliminaries}
% In this section, we formulate our problem setting as the teacher-assisted Markov decision process (MDP), where the learning agent could actively query a teacher for various types of feedbacks during an episode. The overall goal of the agent is to learn efficiently from the teacher's feedback with the minimum number of queries. Below we formalize the problem setting as well as the teacher protocol.

\subsection{Motivation: referentiality, productivity, and pragmatism in human communication}
\label{sec:human-comm}

Humans are well-adapted for progressively efficient communication.
First of all, the human language is \textit{referential} \citep{tomasello2005constructing,sievers2016reference}, containing symbols that can refer to abstract concepts which are detached from the current time and space. 
Second, the human language is highly \textit{productive} \citep{baayen1991productivity,piantadosi2017infinitely}. 
Humans can define new terms by composing the meanings of existing ones. 
They can convey an idea using diverse expressions that manifest different levels of abstraction. 
Finally, human communication is strongly \textit{pragmatic} \citep{wittgenstein1953philosophical,austin1975things,clark1996using,friedpragmatics}.
They have communication objectives and can flexibly adapt their language to meet those objectives. 
The principle of \textit{least collaborative effort} \citep{clark1986referring} posits that human communication is guided by a desire to reduce the joint effort of the interlocutors. 
Our work is inspired by these characteristics of human communication. 
We will show that endowing AI agents with similar characteristics can naturally lead to the emergence of progressively efficient communication.

\subsection{Learning as communication}
\label{sec:learn-as-comm}

We consider a setting where a learner interacts with a teacher to learn to perform a task in an MDP environment with state space $\gS$ and (primitive) action space $\gA$.
Let $\gG$ be the set of all the tasks in the environment that the teacher can recognize and evaluate. 
These tasks are related: some tasks can be subtasks of others (e.g., the nodes of the tree in \autoref{fig:demo}). 
The learner aims to master a task $g \in \gG$, referred to as the \textit{main task}.
At test time, it has to perform this task independently without the guidance of the teacher. 
% The agent performs tasks by defining a policy $\agentpol(s, g)$ that takes as input a state $s$ and a task $g$ and return a distribution over actions.
% The action space of this policy varies depending on the learning framework, which may not be the same as the environment's action space. 

% At time step $t$, the agent is in state $s_t$.
% Taking action $a_t$ transitions it to a new state $s_{t + 1} \sim E(s_t, a_t)$.

In this paper, we view learning as a communicative activity, in which two interlocutors, the learner and the teacher, exchange communication signals in order to align their mental states.
In this case, they want to align their interpretations of the main task. 
The communication signals can be expressed in any kind of language.
To better understand this view, consider the IL setting \citep{ross2011reduction}, where 
the communication signals are made up of the primitive actions of the learner.
In each episode, the learner first takes actions in the environment to perform the main task.
These actions represent how it interprets the task.
The teacher then replies with a sequence of reference actions to correct the learner's interpretation.
Alignment is achieved when the learner presents a task performance that the teacher accepts and does not propose any further adjustment.

\subsection{Limitations of traditional learning frameworks}

Communication in standard (non-hierarchical) IL or RL is inefficient because it lacks referentiality, productivity, and pragmatism.
The learner in these frameworks ``speaks'' by taking primitive actions.
This form of communication is non-referential, as primitive actions do not allude to any abstract concepts. 
It is also unproductive because the primitive-action space stays fixed during the course of learning.
Meanwhile, the teacher in these frameworks also employs a non-referential, static language. 
The IL teacher can only communicate through the learner's primitive actions, whereas the RL teacher can provide only numerical feedback, which, like primitive actions, has limited capability of referring to abstract concepts and composing new meanings.

Without a referential, productive language, sophisticated pragmatism does not emerge.
Since the learner and the teacher speaks with only one level of abstraction, they do not need to reason pragmatically to decide on the most appropriate level of abstraction for expressing an intention.
For example, in the environment illustrated in \autoref{fig:demo}, to convey the intention of ``bake pork'', an IL learner has no other choice but to perform the task in actions. 
In contrast, a human can generate diverse expressions to refer to this task (e.g., ``\textit{bake pork}'', ``\textit{get coal, hit pig}'', ``\textit{go up, down, left...}''), and selects among them an expression that optimizes a communication goal to convey to another human.
\looseness=-1

%% file: sections/overview.tex
\section{Communication-efficient interactive learning (\ouralgo)}

\subsection{Overview}
\label{sec:overview}

Before delving into the implementation details, we first highlight the key novelties of our framework. 

\paragraph{Referential, productive communication with abstract intentions.}
\ouralgo allows the learner and the teacher to convey \textit{abstract intentions}.
An intention is a symbol referring to a sequence of actions that aims to accomplish a task (e.g., ``bake pork'', ``make arrow'').
Using intention as the medium, \ouralgo supports referential communication.
Moreover, the framework allows for the expansion of the set of intentions and the composition of existing intentions for expressing new intentions, making communication also productive.

Adopting the perspective introduced in \autoref{sec:learn-as-comm}, we also frame learning in \ouralgo as a communicative activity, where the goal is for the learner and the teacher to agree on the meaning of the intention referring to the main task. 
To achieve this goal, the two interlocutors repeat a process in which the learner conveys its current interpretation of the intention, and the teacher provides feedback to rectify that interpretation. 
To convey its interpretation of an intention, the learner can select among two general options:
\begin{enumerate}[label=(\alph*), nolistsep]
\item \textbf{verbal communication}: express the intention in terms of other intentions;
\item \textbf{non-verbal communication}: execute the intention by taking actions in the environment to perform the task that it refers to.
\end{enumerate}
For option (a), in our implementation, the learner utters only an initial part of the expression so that teacher can correct it more efficiently.
For example, when learning a ``bake pork'' task, it would say ``\textit{[to bake pork, I will] get coal}'' rather than ``\textit{[to bake pork, I will] get coal, then hit pig}''.

\begin{wrapfigure}{r}{0.33\textwidth}
  \vspace{-.5cm}
  \centering
    \includegraphics[width=0.33\textwidth]{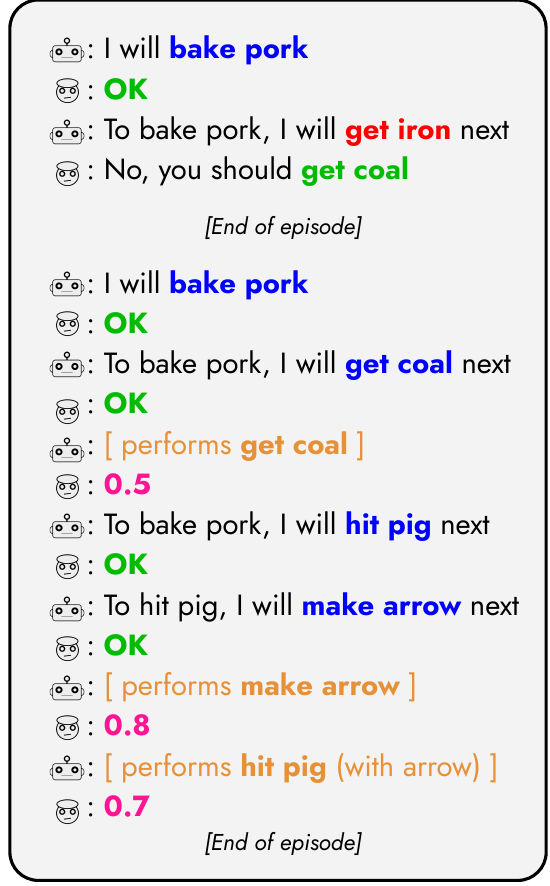}
  \caption{Conversations between the learner and teacher for learning the ``bake pork'' task.}
  \label{fig:conversation}
  \vspace{-.8cm}
\end{wrapfigure}

Upon observing the learner's interpretation, the teacher offers feedback.
\ouralgo can be instantiated with various types of feedback.
In this paper, we present a variant where the teacher issues IL-like \textit{instructive feedback} and RL-like \textit{evaluative feedback}.
Specifically, if the learner chooses option (a), the teacher provides instructive feedback in the form of intention correction.  
When the proposed intention is correct, the teacher simply confirms.
If the learner chooses option (b), the teacher offers evaluative feedback in the form of numerical evaluation of task execution.
With this option, other types of feedback like language descriptions \citep{nguyen2021interactive} can be incorporated.
We choose numerical feedback to simplify the learning algorithm. 

\autoref{fig:conversation} shows example conversations between the learner and the teacher.
Since our focus is on communication at the intention level, we simplify the problems of generating natural expressions to convey intentions and interpreting intentions from natural expressions. 
Despite the uncomplicated look of the language, our communication protocol highly general, as the learner and the teacher can potentially exchange a broad set of intentions. 
Particularly, the protocol strictly generalizes those of IL and RL: (hierarchical) IL is equivalent to \ouralgo with only verbal communication, while RL (with sparse reward) is analogous to \ouralgo with only non-verbal communication.
\looseness=-1

\paragraph{Pragmatic communication to save long-term effort.} Within the realm of verbal communication, \ouralgo allows the interlocutors to convey expressions of varying levels of abstraction.
We adopt a simple notion of level of abstraction, defining it as the number of actions an intention refers to. 
For example, in \ouralgo, the learner may utter either ``\textit{[to bake pork I will] make stone pickaxe}'' or ``\textit{[to bake pork I will] get stone}'' to refer to an initial step of ``bake pork'', but the former is a more abstract expression because ``get stone'' is a subtask of ``make stone pickaxe''.

Having flexibility in the language enables the learner to communicate pragmatically to optimize for a goal.
In \ouralgo, the learner's goal is to minimize  long-term (joint) communication effort.
We express this goal by setting a learning objective that minimizes short-term communication effort while also minimizing task error.
Striving for this objective drives the learner to speak with the teacher at a level of abstraction that best suits their current mutual knowledge, and to gradually speak more abstractly.
Specifically, the learner begins by communicating using the least abstract intentions, as the meanings of those are easiest to learn.
At this stage, the communication effort is substantial: the learner proposes numerous intentions during an episode, inducing tantamount effort from the teacher to provide feedback.
\ouralgo allows the learner to explore, occasionally uttering more abstract intentions.
The urge to save communication effort prompts the learner to improve its understanding of these abstract intentions (by incorporating the teacher's feedback), and to leverage them incrementally more often to shorten its verbal expressions. 
When the learner has mastered the main task, it will communicate minimally with the teacher.
It will simply declare the intention of performing the task (e.g., ``\textit{[I will] bake pork}'') and execute that intention immediately, which is the same as its behavior at test time.
\looseness=-1

It is important to emphasize that \ouralgo does \textit{not} directly force the learner to speak increasingly more abstractly.
Much like in humans, this capability emerges as a means for the learner to achieve its \textit{socially motivated goal}---to communicate effectively with minimal effort.

\paragraph{Pragmatic teachers.} Another novelty of our framework is the employment of teachers modeled after human pragmatic behavior. 
Hierarchical IL or RL (e.g., \citep{le2018hierarchical}) typically assumes a \textit{top-down} teacher, who always recommends the most abstract intention to correct the learner's intention. 
Consider the example in \autoref{fig:conversation}, where the learner falsely proclaims ``\textit{[to bake pork I will] get iron}''. 
According to the task tree in \autoref{fig:demo}, ``get coal'', ``make stone pickaxe'', ``get stone'' are all valid intentions to refer to the learner. 
A top-down teacher would choose ``get coal'', the most abstract one, to utter.
This teacher is non-pragmatic because it ignores the learner's behavior. 
To better mimic communication with humans, we simulate two types of pragmatic teacher, each of which employs a heuristic to select an intention that is deemed easiest for the learner to interpret.
The \textit{language-based} teacher replies with the intention whose level of abstraction is most similar to that of the learner's proposed intention. 
Meanwhile, the \textit{performance-based} teacher samples an intention among the candidates with probability proportional to the historical success rate of the learner on executing it. 
More details about these teachers are given in \autoref{sec:prag-teacher}. 

Unlike the top-down teacher, the two pragmatic teachers lack motivation to communicate abstractly. 
We will empirically demonstrate that regardless of whether the teacher possesses this motivation, our learner is still able to drive communication to be incrementally more abstract and efficient.

%% file: sections/implementation.tex
\subsection{Enriching the means: communication of referential intentions}
\label{sec:enrich_means}

\paragraph{Intentions.} 
Let $\gI \subseteq \gG$ be the set of intentions the learner can convey.
Initially, it contains a set of seed intentions, which includes the intention of performing the main task. 
The seed intentions can be constructed with a knowledge base. 
For example, one can collect a comprehensive list of MineCraft tasks from the game's Wiki. 
Moreover, this set is \textit{expandable}: when the teacher introduces a new task to the learner, it can add to set the task's name as a new intention.
We do not specify the levels of abstractions of the intentions to the learner. 
It realizes this quality through interaction with the teacher. 
\looseness=-1

\begin{wrapfigure}{R}{0.48\textwidth}
    \vspace{-0.8cm}
    \begin{minipage}{0.48\textwidth}
      \begin{algorithm}[H]
        \small
        \caption{\ouralgo training episode}
        \label{alg:ceil}
        \begin{algorithmic}[1]
          \STATE Observe initial state $s_1$ and main task $g$
          \STATE Set initial intention $i_1 = g$
          \STATE Initialize memory stack $M = \{ i_1 \}$
          \STATE $t = 0$
          \WHILE{not $M\textrm{.empty()}$}
          \STATE $t \leftarrow t + 1$
          \STATE Get current intention $i_t = M\textrm{.top()}$
          \STATE Choose action $u_t \sim \agentpol(s_t, i_t)$
          \IF[verbal]{$u_t \in \gI \cup \{ \texttt{\doneact}\}$} 
          \STATE Receive instructive feedback $f_t = u^{\star}_t$
          \STATE Stay in the same state $s_{t + 1} = s_t$
          \STATE Set new intention $i_{t + 1} = M(u_t)$
          \ELSE[non-verbal]
          \STATE Execute $i_t$ and arrive in final state $s^L_t$
          \STATE Receive evaluative feedback $f_t \in \mathbb{R}$
          \STATE Set new state $s_{t + 1} = s^L_t$
          \STATE Set new intention $i_{t + 1} = M(\doneact)$
          \ENDIF
          \ENDWHILE
          \STATE Update policy $\agentpol$ w.r.t. learning objective
        \end{algorithmic}
      \end{algorithm}
    \end{minipage}
\vspace{-0.5cm}
\end{wrapfigure}

\paragraph{Learner components.} The learner has two components: a \textit{policy} $\agentpol(u \mid s, i)$ and a \textit{memory} $M(u)$.
The policy takes as input an environment state $s \in \gS$ and an intention $i \in \gI$, and outputs a distribution over actions $u \in \gI \cup \{ \doact, \doneact \}$, where $\doact$ and $\doneact$ are  special actions which we will define shortly.
The memory helps the learner update and keep track of its current intention. 
Each action $u$ can either be verbal or executive.
Taking a verbal action ($u \in \gI \cup \{ \doneact \}$) alters the learner's current intention (its mental state), whereas taking an executive action ($u = \doact$) changes its environment state. 

\paragraph{Interactions during an episode (Algorithm \ref{alg:ceil}).} 
The learner starts in state $s_1$ and conveys the intention $i_1 = g$, which is to perform the main task. 
At the time step $t$, suppose the learner's intention and environment state are $i_t$ and $s_t$, respectively. 
The learner selects an action $u_t \sim \agentpol(s_t, i_t)$.
If $u_t \in \gI \cup \{ \doneact \}$, the learner chooses to verbally communicate an intention. 
The $\doneact$ action represents the intention of relinquishing the current intention.
The learner computes a new current intention $i_{t + 1} = M(u_t)$ by querying the memory with the action.
Meanwhile, the environment state stays the same, $s_{t + 1} = s_t$. 
After that, the learner receives instructive feedback $f_t = u^{\star}_t \in \gI \cup \{ \doneact \}$ from the teacher, which indicates the correct intention in the current state.  

If $u_t = \doact$, the learner elects to execute the current intention $i_t$.
In this case, the agent continuously taking primitive actions until it decides to terminate, generating an execution $(s_t^{1} = s_t, a_t^1, \cdots, s_{t}^L, a_{t}^L)$, where $L$ is the trajectory length, $a_{t}^l \in \gA \text{ for } 1 \leq l \leq L$. 
It then updates its current intention and environment state, setting $i_{t + 1} = M(\doneact)$ and $s_{ t+1} = s_t^L$, and receives evaluative feedback from teacher, which is a score $f_t \in \mathbb{R}$ judging the execution. \looseness=-1

\paragraph{Memory.} Following \citet{nguyen2022framework}, we implement the memory $M$ as a stack data structure, which prioritizes the most recently proposed intentions. 
Initially, the stack contains only the main-task intention $i_1$.
When the learner chooses to communicate verbally, the communicated intention is pushed to the stack. 
However, when the special \texttt{\doneact} action is taken, the intention at the top of the stack is popped.
After pushing or popping, the intention at the top of the stack is returned as the current intention.
If there is no intention in the stack to return, the episode ends. 
This stack-based memory allows the learner to learn a deep hierarchy of tasks, making \ouralgo more general than prior work on hierarchical IL and RL that assumes only a two-level hierarchy \citep{kulkarni2016hierarchical,le2018hierarchical}.
We defer the exploration of more intricate memory designs, such as interleaving executions of multiple tasks or prioritizing tasks that cost less resources to execute from the current state. 
\looseness=-1

\subsection{Injecting the motivation: minimization of long-term communication effort}
\label{sec:inject_motivation}

\paragraph{Learner's intrinsic motivation.} 
As mentioned, the \ouralgo learner aims to minimize the long-term communication effort. 
To formalize this goal, let $\pi_0$ be the learner's current policy and $\pi_n$ be the its policy after $n$ learning updates.
For simplicity, we assume the policy is updated after every learning episode.
We associate a communication cost $c(s, i, u)$ with every action $u$ taken by the learner when it is in state $s$ with intention $i$. 
$C(\tau) = \sum_{t = 1}^T c(s_t, i_t, u_t)$ represents the communication effort in an episode an episode $\tau = \{ (s_t, i_t, u_t) \}_{t = 1}^T$. 
Let the \textit{task error} $J_{\textrm{err}}(\pi)$ be a function that quantifies the degree of misalignment of a policy $\pi$ with respect to the teacher's expectation, where
$J_{\textrm{err}} = 0$ indicates perfect alignment. 
In our setting, $J_{\textrm{err}}(\pi)$ reflects the average number of incorrect intentions proposed by $\pi$, and the negative average score of its intention executions.
We define the number of learning episodes $N$ as the smallest integer $n$ such that $J_{\textrm{err}}(\pi_n) = 0$.

The \textit{long-term communication effort} is defined as the communication effort accumulated across all future learning episodes
\begin{align}
    J(\pi_1) = \E_{\tau_1 \sim \pi_1, \cdots, \tau_{N} \sim \pi_{N}} \left[ \sum_{n = 1}^{N} C(\tau_n) \right]
\end{align} where $\tau \sim \pi$ denotes generating a learning episode with policy $\pi$, and $\pi_n = \texttt{Improve}(\pi_{n- 1}, J)$ with \texttt{Improve} being an optimizer (e.g., Adam) that computes a new policy $\pi_n$ such that $J(\pi_n) < J(\pi_{n - 1})$.  
The expectation is taken over all possible sequences of future episodes. 

Computing this objective is impractical, so we resort to an approximation scheme. 
We split $J(\pi_1)$ into two terms, $\mathbb{E}[C(\tau_1)]$ and $\mathbb{E}[C(\tau_2) + \cdots C(\tau_{N})]$. 
The first term, $\mathbb{E}[C(\tau_1)] \triangleq J_{\textrm{com}}(\pi_1)$, represents the communication effort in the next episode and can be effectively optimized with an RL algorithm. 
While we cannot directly optimize the second term, we aim to minimize the number of terms in the summation, i.e. minimizing $N$. 
To do that, we heuristically minimize $J_{\textrm{err}}(\pi_1)$.
The intuition here is that the less misaligned a policy is, the less learning episodes is needed to perfectly align it.
This may not always be true but we find the heuristic works sufficiently well in practice. 
In the end, the current policy $\pi_0$ is updated to a new policy $\pi_1$ that satisfies $J_{\textrm{com}}(\pi_1) + J_{\textrm{err}}(\pi_1) < J_{\textrm{com}}(\pi_0) + J_{\textrm{err}}(\pi_0)$.
 
Both terms in the new objective are essential. 
If trained to minimize only $J_{\textrm{com}}(\pi_1)$, the learner would quickly resort to taking only a single \doact action during an episode. 
This behavior resembles a sparse-reward RL setting, in which the learner executes the main task and receives a single reward.
In this case, because sparsely provided rewards are weak learning signals, the learner would need more learning episodes to master the task, ending up incur more long-term communication effort. 
On other hand, if solely aiming to reduce $J_{\textrm{err}}(\pi_1)$, the learner would not be strongly motivated to attempt communication at a higher level of abstraction.
It may be content with proposing and executing low-level intentions because those are easy to execute accurately. 

\paragraph{Learning algorithm.} 
We propose an algorithm that extends Q-learning to optimize for the learner's objective. 
We define the reward function $r(s, u; i) = -c(s, i, u)$ and the optimal Q-function $Q^{\star}(s, u; i)$ based on $r$.
We approximate this function by $Q_{\theta}(s, u; i)$ and define the learner's policy as $\agentpol(u \mid s, i) = \mathds{1}\{u = \argmax_{u'} Q_{\theta}(s, u'; i)\}$.
In each episode, we generate a trajectory using the current policy and store it in a replay buffer. 
Then we sample a batch of transitions $\{ (s_t, i_t, u_t, r_t, s_{t + 1}, i_{t + 1}) \}_{i = 1}^B$ from the buffer to update the Q-function. 
To minimize the next-episode communication effort, we apply the standard Q-learning update:
\begin{align}
    \theta_{\textrm{new}} = \min_{\theta} \frac{1}{B} \sum_{i = 1}^B \left(Q(s_t, u_t; i_t) - (r_t + \gamma \max_{u} Q_{\theta}(s_{t + 1},  u; i_{t + 1})) \right)^2
\label{eqn:rl_update}
\end{align} where $\gamma$ is a discount factor. 
To reduce the task error, we enables the learner to improve by incorporating the teacher's feedback.
We consolidate the instructive feedback using a max-margin objective:\looseness=-1 %by CITE: 
\begin{align}
    \theta_{\textrm{new}} = \min_{\theta} \frac{1}{B} \sum_{i = 1}^B \max(0, \lambda + \max_{u \neq \{ \doact, u^{\star}\}} Q(s_t, u; i_t) - Q(s_t, u_t^{\star}; i_t)) 
\label{eqn:il_update}
\end{align} which aims to separate the Q-value of the correct intention $u^{\star}_t$ from others by a margin of at least $\lambda$.
To integrate the evaluative feedback, we implement a weighted self-imitation learning approach, which first computes the  objective in \autoref{eqn:il_update} with $u^{\star}_t$ being the primitive actions taken during an execution, and weights this objective by the numerical score provided by the teacher.

%% file: sections/exp_setup.tex
\section{Experimental Setup}
\label{s5:exp_setup}
\textbf{MineCraft environment.} We employ the simulator developed by \citet{sohn2018hierarchical}, which emulates a MineCraft-style game on a 2D grid. 
The player can traverse in an environment, interact with other entities to collect items, and combine them to create new items. 
The state given to the agent is a $10 \times 10 \times 19$ grid with 19 channels representing the locations of the agent and other entities, and the terrain.
The full task hierarchy is given in \autoref{sec:task-graph}. 

\textbf{Learner's policy.} The Q-function of the learner is a CNN-based neural network that maps a state tensor and a one-hot vector representing an intention to a distribution over actions. 
The detail architecture is given in \autoref{sec:impl}.

\textbf{Baselines.} We compare with the followings: (a) \textit{flat imitation learning} (FIL) implements DAgger \citep{ross2011reduction}, (b) \textit{flat reinforcement learning} (FRL) is standard Q-learning, (c) \textit{hierarchical imitation learning} (HIL) performs DAgger with a learner that proposes high-level intentions and (d) \textit{active hierarchical imitation learning} (AHIL) learns a success predictor to determine when to execute an intention. 
The labels for training this predictor are determined in hindsight.
Among these baselines, FIL and FRL cannot communicate at a high level of abstraction. 
HIL can do so but lacks a motivation for minimize communication effort. 
AHIL can be viewed as an ablated version of \ouralgo that is only motivated to minimize learning error ($J_{\textrm{err}}$).

\textbf{Training settings.} We evaluate all approaches on three settings. 
In the \textit{learn-from-scratch} setting, for every method, we train a randomly initialized agent to perform the BakePork task, which requires on average 68 actions to complete. 
In the \textit{environment-adaptation} setting, we put the learned agents in new environment layouts and continue training them on the BakePork task.
In the \textit{task-adaptation} setting, we instead train these agents to perform two new tasks, BakeBeef and SmeltSilver, which share several common subtasks with the BakePork task.
Each agent is given a budget of 3M feedback requests in the learn-from-scratch setting and 10K in the two adaptation settings. 

\textbf{Teacher models.} We employ the three teacher models described in \autoref{sec:overview}.

%% file: sections/results.tex
\section{Results}
\label{s6:results}
\begin{figure}[htbp!]
\vspace{-0.5em}
\begin{subfigure}{1.0\textwidth}
         \centering
         \includegraphics[scale=0.35]{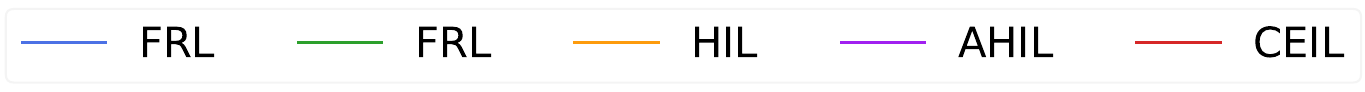}
     \end{subfigure}
    \\
    \centering
     \begin{subfigure}{0.3\textwidth}
         \centering
         \includegraphics[width=0.9\textwidth]{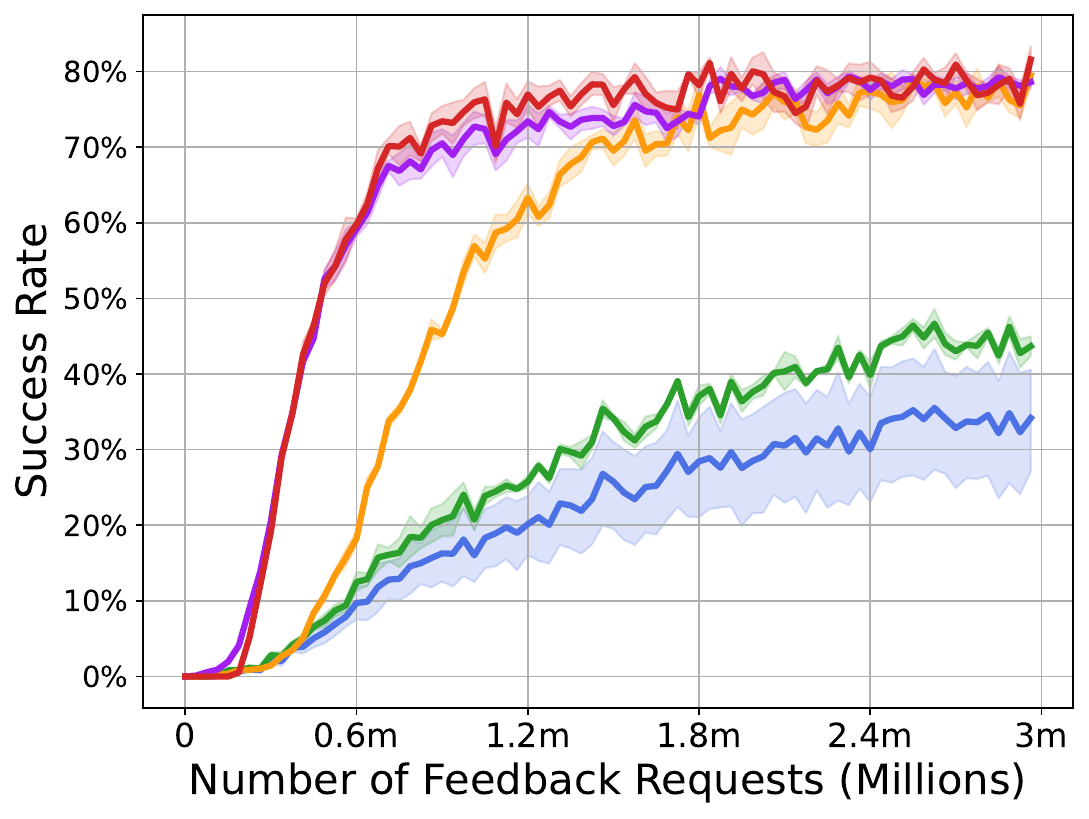}
         \caption{Top-down (non-pragmatic) teacher}
     \end{subfigure}
     \hfill
     \begin{subfigure}{0.3\textwidth}
         \centering
         \includegraphics[width=0.9\textwidth]{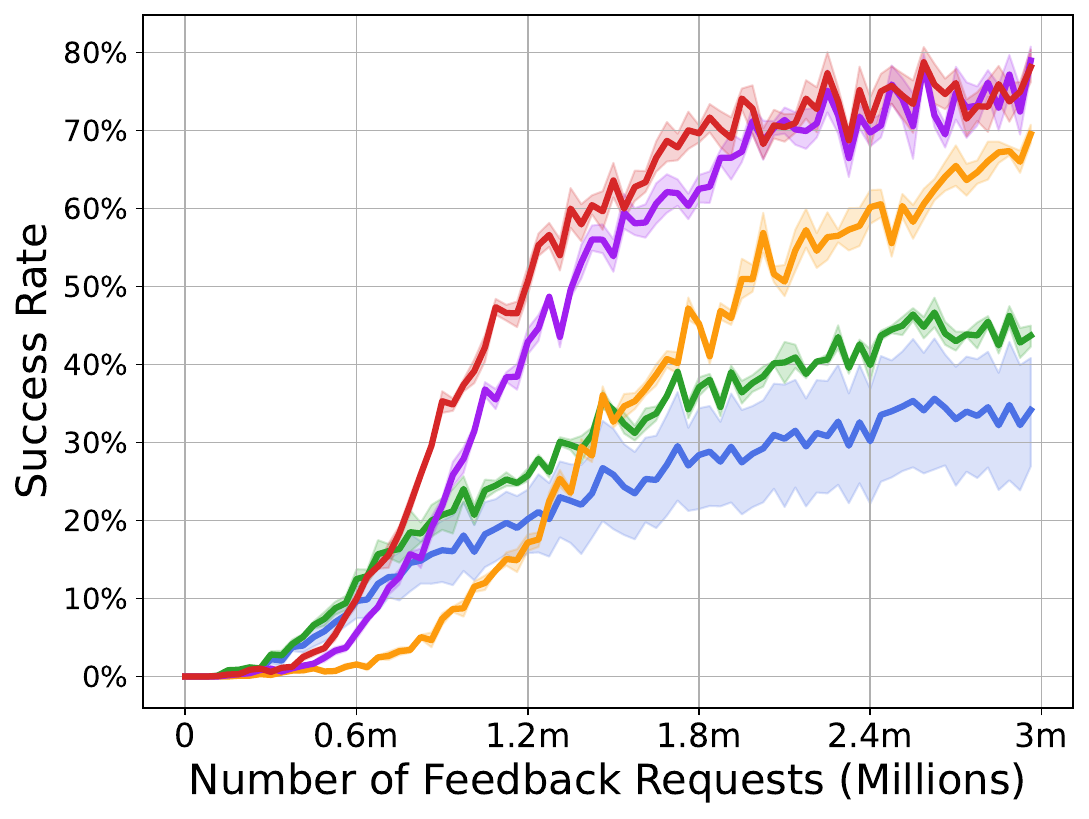}
         \caption{Language-based pragmatic teacher}
     \end{subfigure}
     \hfill
     \begin{subfigure}{0.3\textwidth}
         \centering
         \includegraphics[width=0.9\textwidth]{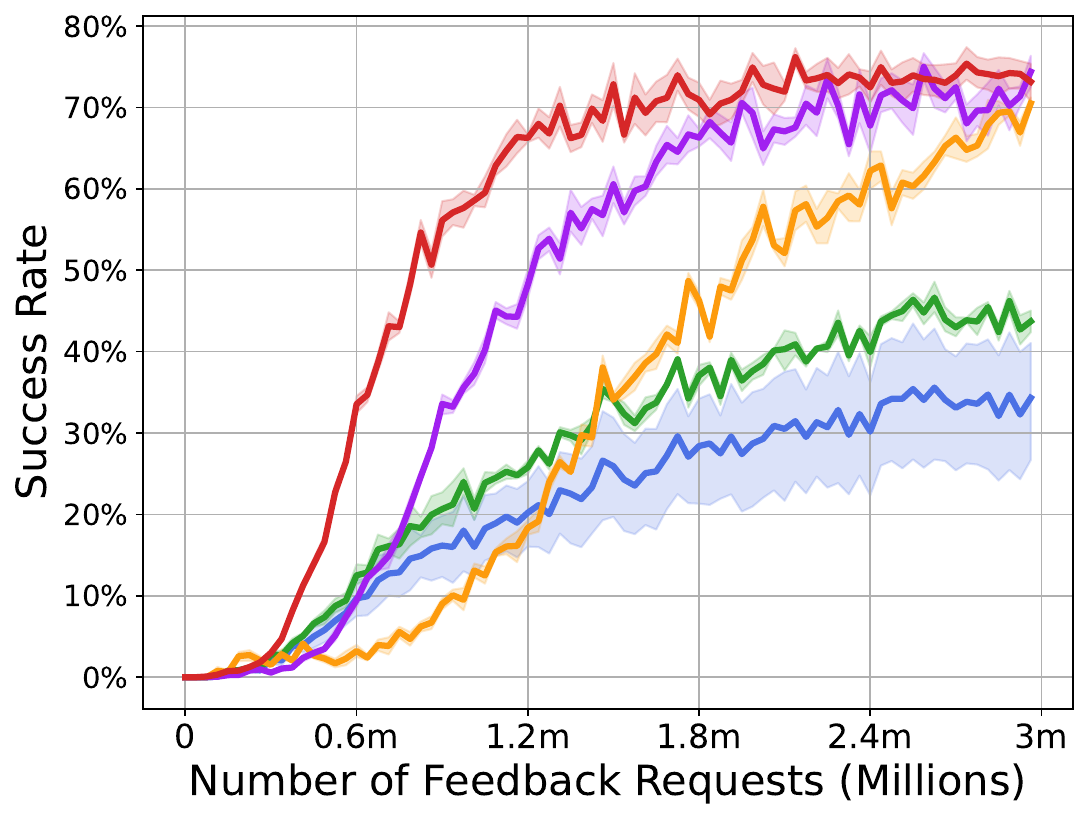}
         \caption{Performance-based pragmatic teacher}
     \end{subfigure}
\caption{Success rate on the training task (BakePork) as a function of number of feedback requests made to the teacher. Results are averaged over four random seeds. }
\vspace{-0.5em}
\label{fig:train-task-perf}
\end{figure}

\paragraph{Learn-from-scratch setting.}
In \autoref{fig:train-task-perf}, we plot the success rate on unseen environment layouts against the number of feedback requests for all teacher models. 
The flat IL and RL baselines, which communicate only through low-level media, cannot reach a high success rate even after 3M feedback requests. 
\ouralgo consistently outperforms all baselines in terms of both final success rate and communication efficiency.
With the non-pragmatic top-down teacher, \ouralgo reaches similar performance as HIL but it learns much faster. 
For example, to reach a success rate of 70\%, \ouralgo requires just over 600K requests while HIL needs 1.5M.
With the pragmatic teachers, \ouralgo not only learns faster but achieves a higher final success rate than HIL (e.g., +10\% on the setting with the performance-based teacher). 
\ouralgo improves remarkably faster than AHIL when paired with the performance-based teacher, showing the necessity of minimizing the short-term communication effort ($J_{\textrm{com}}$). 
Overall, these results showcase the effectiveness of \ouralgo against diverse teacher behaviors, especially those that resemble human pragmatic behavior.

\begin{figure}[htbp!]
    \hspace{1.5em}
    \begin{subfigure}{0.45\textwidth}
         \centering
         \includegraphics[width=0.7\textwidth]{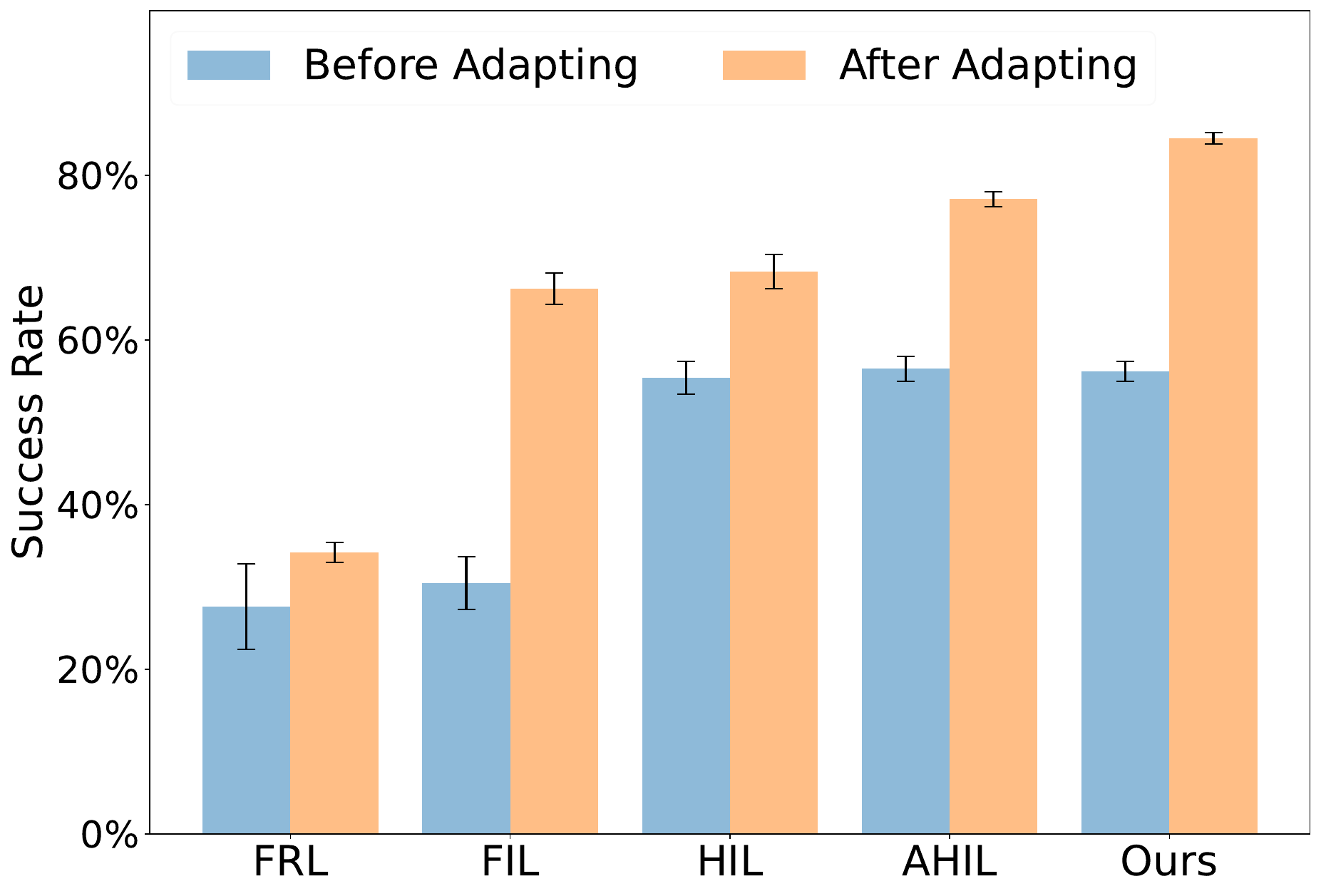}
         \caption{Adaptation to new environment layouts}
         \label{sfig:adapt_env}
     \end{subfigure}
     \begin{subfigure}{0.45\textwidth}
         \centering
         \includegraphics[width=0.7\textwidth]{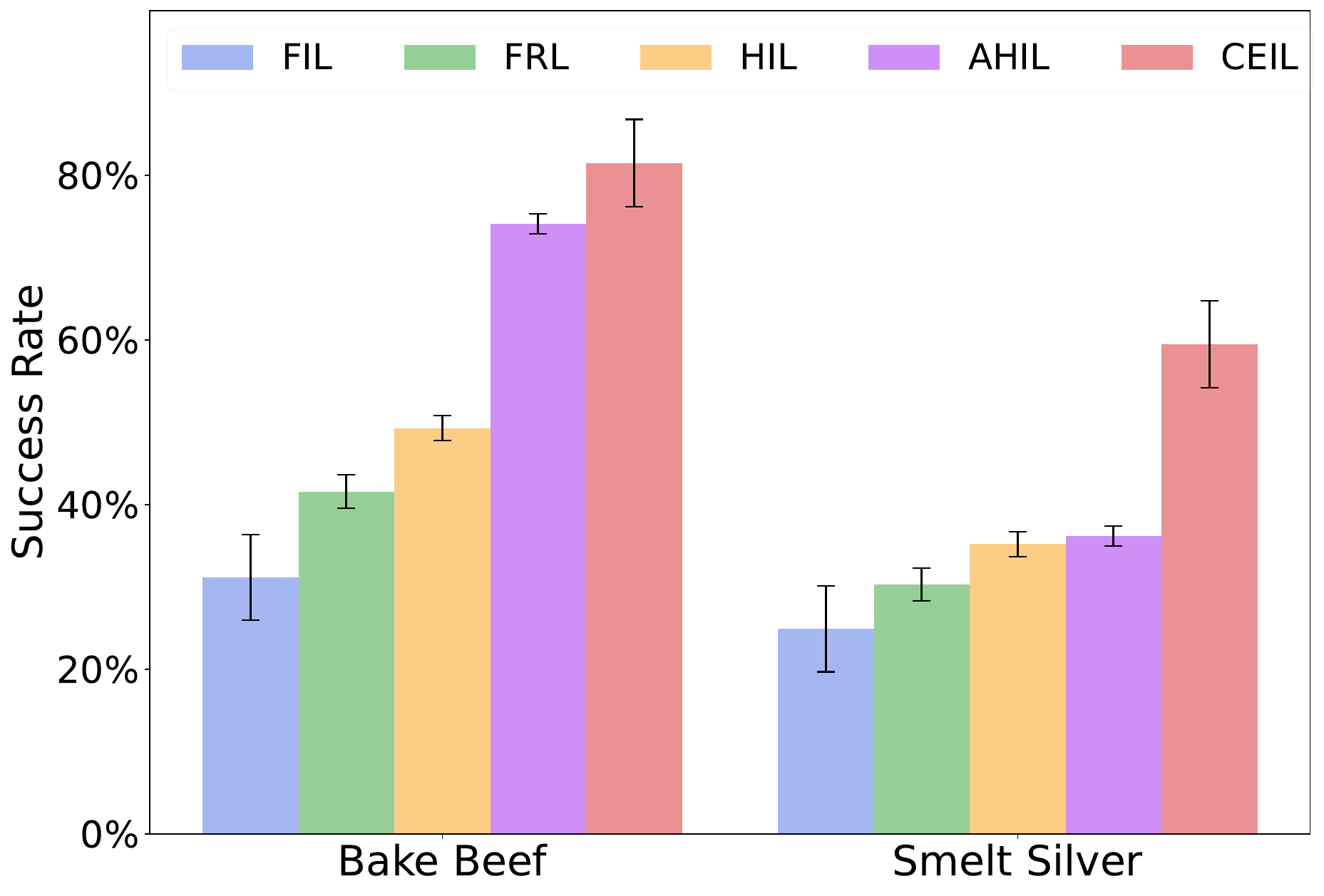}
         \caption{Adaptation to new tasks}
         \label{sfig:adapt_skill}
     \end{subfigure}
     \hfill
\caption{Performance of different approaches when adapted to new BakePork environments and to new tasks (BakeBeef and SmeltSilver). All agents learn with the performance-based pragmatic teacher. They were pre-trained on the BakePork task to reach the same success rate, and then are adapted with the same request budget of 10K. In both settings, \ouralgo adapts more successfully, showing that the mutual knowledge it acquired with the teacher during the learn-from-scratch phase generalizes more robustly.}
\vspace{-1.0em}
\label{fig:finetune-task-perf}
\end{figure}

\paragraph{Adaptation settings.} 
\autoref{fig:finetune-task-perf} shows the performance of all approaches on the two adaptation settings. 
\ouralgo attains the highest final success rate in all settings. 
Even though \ouralgo, HIL, and AHIL were pre-trained to have the same success rate on the BakePork task, their post-adaptation capabilities differ significantly. 
Most vividly, in the SmeltSilver task,  \ouralgo outperforms HIL and AHIL by a subtantial margin of more than 20\% in absolute success rate.  
The gaps in the BakeBeef and BakePork adaptation settings are smaller, possibly because the generalization challenge in these settings is not as significant as in the SmeltSilver setting.
In these tasks, an agent needs to further improve mostly previously learned skills, whereas it has to learn many novel skills to smelt silver. 
These results suggest that \ouralgo not only induces efficient learning but also enables better generalization, which is somewhat surprising because the intrinsic motivation of the \ouralgo learner does not directly aim for the latter.

\begin{figure}[htbp!]
    \centering
     \begin{subfigure}{0.495\textwidth}
         \centering
         \hspace{0.5em}
         \includegraphics[width=0.6\textwidth]{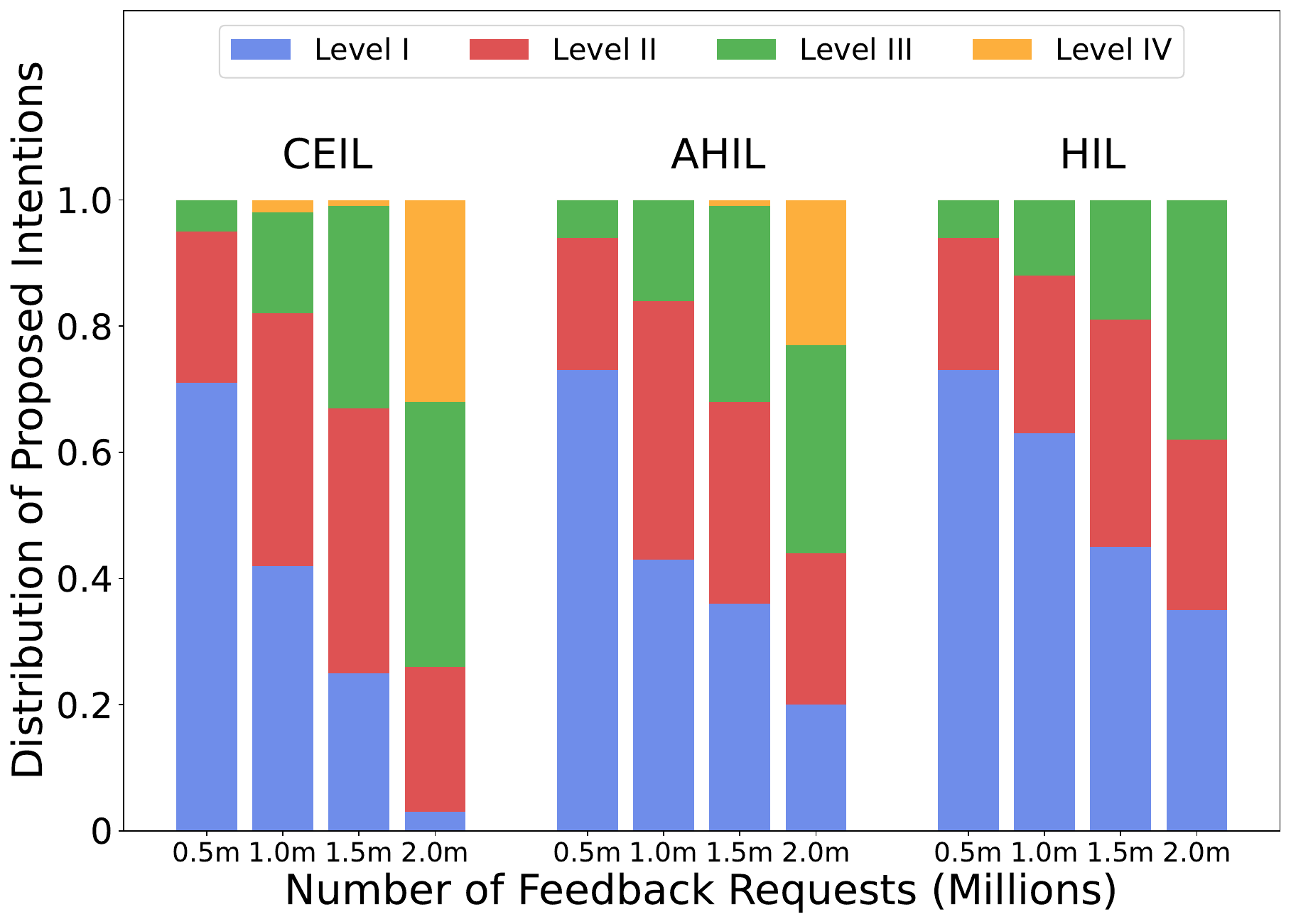}
         \caption{Language-based pragmatic teacher}
          \label{language_oracle_dist}
     \end{subfigure}
     \hfill
     \begin{subfigure}{0.495\textwidth}
         \centering
         \includegraphics[width=0.6\textwidth]{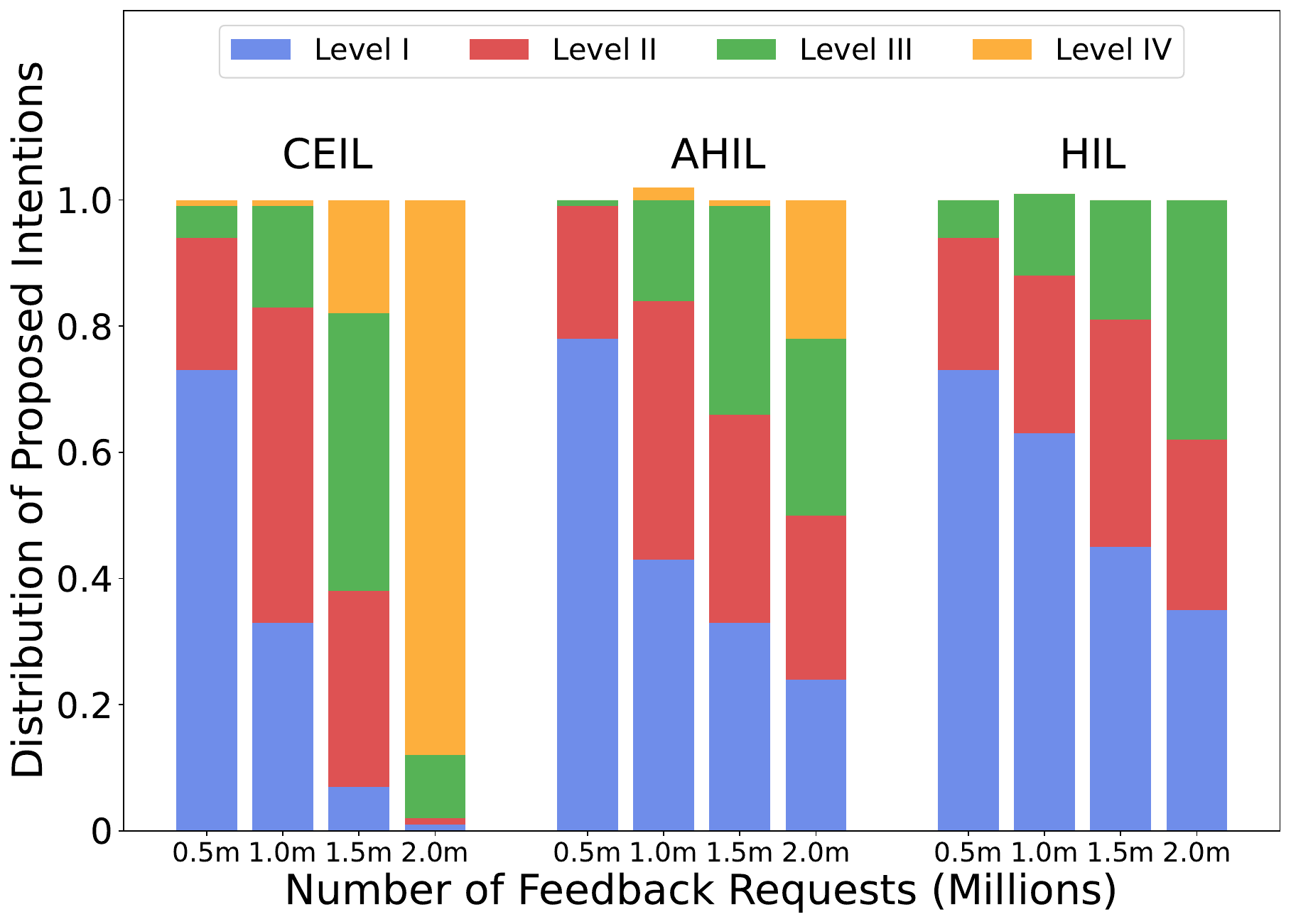}
         \caption{Performance-based pragmatic teacher}
         \label{perf_oracle_dist}
     \end{subfigure}
     \hfill
\vspace{-1.5em}
\caption{Distribution of intentions uttered by an agent over time. We group the intentions into four levels of abstraction, which is a function of the distance to the root in the task tree (level IV is most abstract). \ouralgo exhibits a strong inclination to propose increasingly more abstract intentions.\looseness=-1} 
\label{fig:subtask-dist}
\end{figure}

\paragraph{Does the \ouralgo learner communicate more abstractly over time?}
To answer this question, we visualize the distribution of intentions proposed by an agent over time (\autoref{fig:subtask-dist}).
We divide the intentions into four groups, from least (I) to most abstract (IV).
First of all, we observe that besides \ouralgo, AHIL and HIL also enables the agent to speak more abstractly over time. 
However, our approach induces the strongest exhibition of this phenomenon. 
After 2M feedback requests, \ouralgo shows the largest fractions of the most abstract intention group among the approaches.
Especially, with the performance-based teacher, almost 90\% of intentions are most abstract, where as AHIL only uses them 20\% of the time. 
HIL does not utter the most abstract intentions at all. 
These results confirms our hypothesis that progressively efficient communication emerges when equipping an agent with human-like communication capabilities, specifically a referential, productive language and a desire for least communication effort. 

%% file: sections/related.tex
\section{Related work} 
Previous work on curriculum-based IL and RL \citep{he2012imitation, narvekar2020curriculum, kumar2019reinforcement,zhao2020reinforced,florensa2017reverse,liu2021curriculum} has extended the pragmatic communication capabilities of the teacher by enabling them to choose the order of tasks to teach the learner. 
These strategies can be viewed as making the teacher more pragmatic in choosing \textit{which} intentions to convey.
But in choosing \textit{how} to express those intentions, the teacher is restricted to a low-level language.
Work on hierarchical IL and RL \citep{kulkarni2016hierarchical,sutton1999between,le2018hierarchical} attempts to enrich the language of the teacher and the learner, allowing them to exchange high-level intentions. 
This provides a better means for efficient communication, but the learner still follows the teacher's way of communication without having its own intrinsic motivation to communicate more efficiently over time. 
Moreover, most formulations adopt a hierarchy of only two levels, constraining the flexibility of the language. 

Active learning \citep{ren2021survey, settles2009active} aims to reduce the feedback queries made to the teacher. 
This framework has been instantiated in non-hierarchical IL and RL settings \citep{hsu2019new,brantley2020active,judah2012active,torrey2013teaching}. 
Our work shows that combining active learning with a rich, flexible language results in much more efficient learning. 
Previous active learning strategies are based on intrinsic uncertainty \citep{da2020uncertainty,culotta2005reducing,nguyen2019vision}, error prediction \citep{zhang2016query,nguyen2019help}, or direct optimization of the number of queries via reinforcement learning \citep{fang2017learning,nguyen2022framework}.
We demonstrate that minimizing both the number of queries and the task error is important for progressively efficient communication to emerge strongly.

Recent advances in large language models allow humans to teach them via highly natural language \citep{brown2020language,chowdhery2022palm,OpenAI2023GPT4TR}. 
Users can alternate the behavior of these systems using complex instructions and in-context examples \citep{bubeck2023sparks,wei2022chain}. 
However, due to lacking theoretical guarantees, it remains unclear how to effectively inject intrinsic motivation into these models.
Our work focuses on learning through parameter optimization, which allows us to easily adapt the model continually and enforce intrinsic motivation. \looseness=-1

%% file: sections/conclusion.tex
\section{Conclusion}

In this work, we show that human communication can inspire the next generation of more efficient and robust learning paradigms.
We present an algorithmic framework that emulates several aspects of human communication and show that it outperforms frameworks that lack them.
We hope that our work can inspire new foundations for interactive learning that go beyond the primitive forms of communication employed by IL and RL. 
An interesting future direction to evaluate our framework with real humans or a simulated teacher that can express intentions in a more realistic language.
Incorporating intrinsic motivation into large language models is another exciting direction. 
Recent work \citep{wang2023voyager} has shown that these models can automatically compose increasingly complex skills. 
Equipping them with the ability to continually streamline their communicative and cognitive effort could potentially give rise to more efficient and robust behaviors.

%% file: appendix/a1_subtask_graph.tex
\section{Task graph}
\label{sec:task-graph}

Below is the complete task graph of the MineCraft environment. 
Each node represents a task. Each arrow points from a task to its parents, which it is a subtask of. 
Orange nodes are subtasks of the BakePork task.
Green nodes represent subtasks of the two tasks, BakeBeef and SmeltSilver, which the agents learn during the adaptation settings.
\begin{figure}[h!]
    \centering
    \includegraphics[scale=0.4]{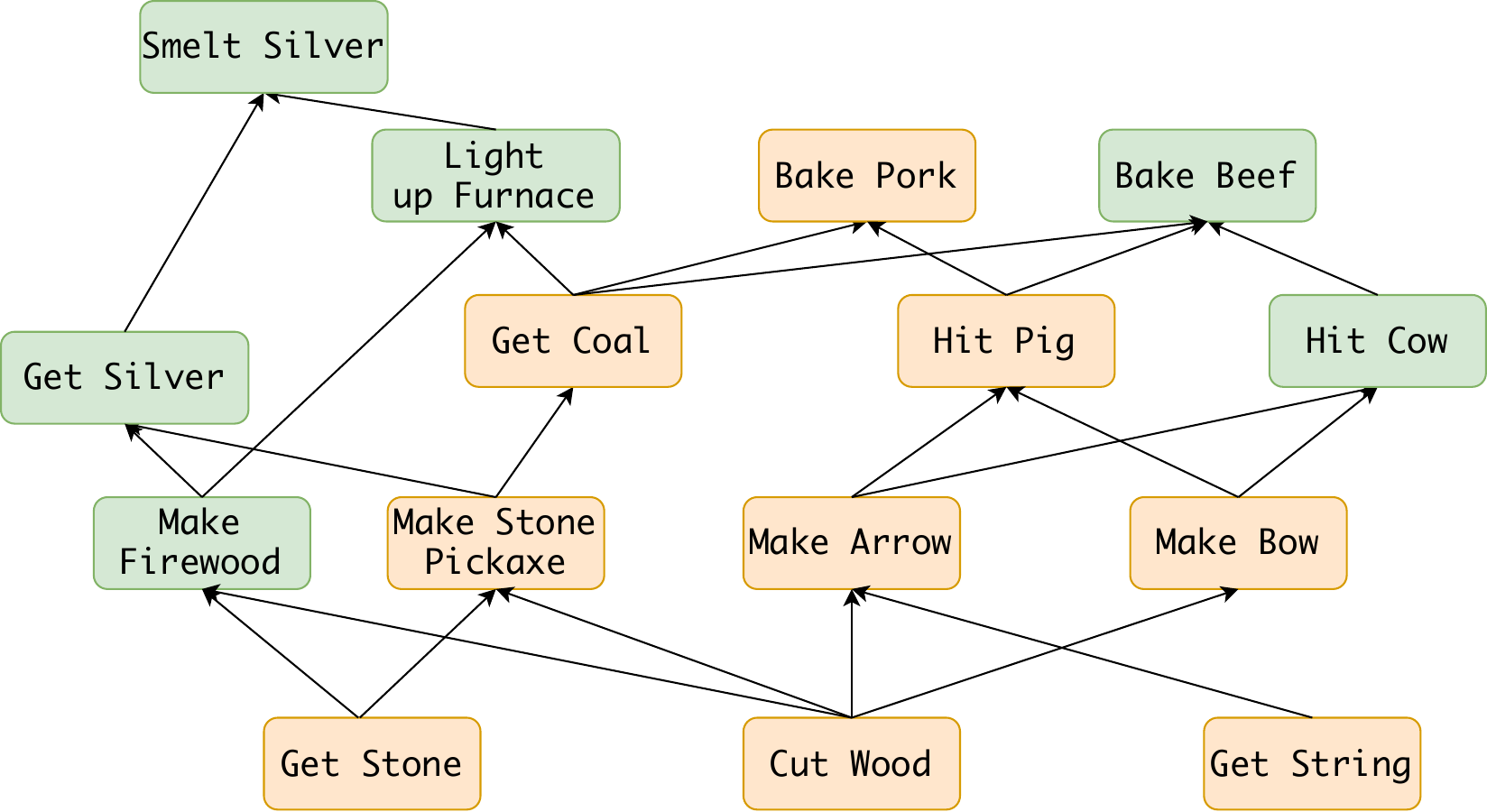}
    % \caption{The task graph of our 2D MineCraft environment. E}
\end{figure}

%% file: appendix/a2_ablation_cost.tex
\section{\ouralgo without minimizing communication cost}

\begin{figure}[htbp!]
\vspace{-0.5em}
\begin{subfigure}{1.0\textwidth}
         \centering
         \includegraphics[scale=0.35]{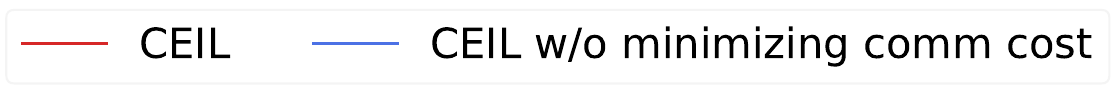}
     \end{subfigure}
    \\
    \centering
     \begin{subfigure}{0.3\textwidth}
         \centering
         \includegraphics[scale=0.23]{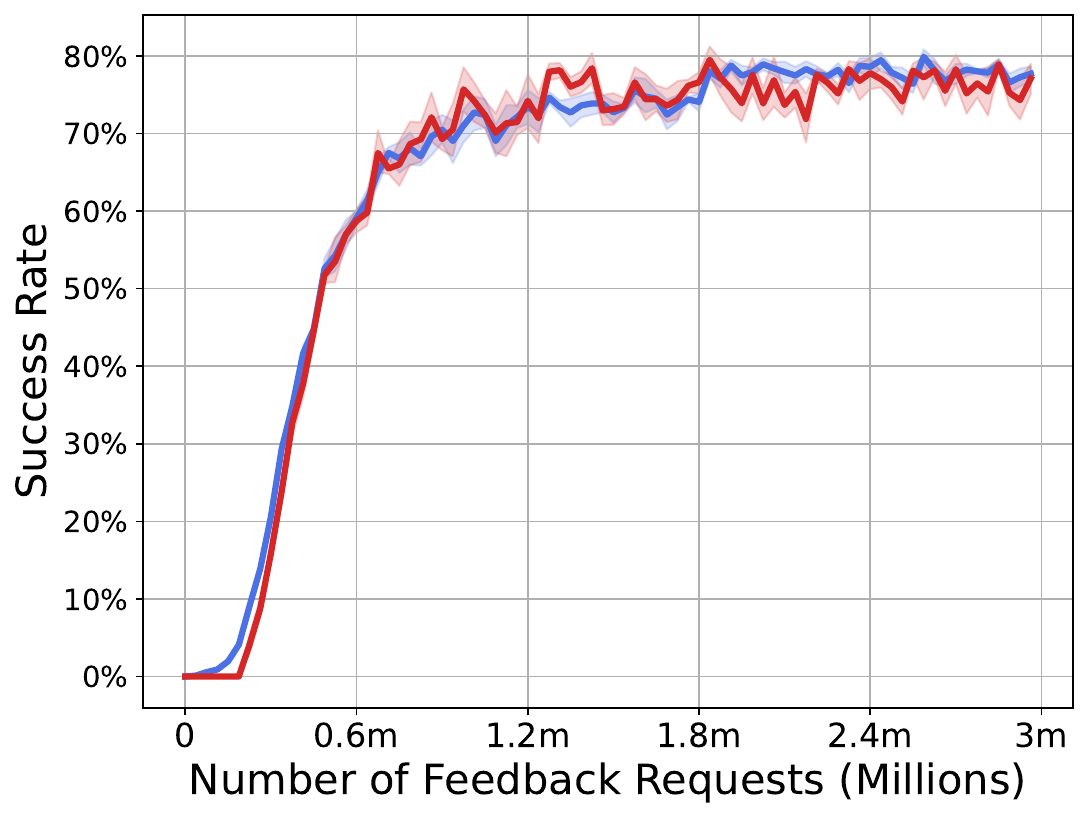}
         \caption{Top-down (non-pragmatic) teacher}
         \label{fig:topdown_comm}
     \end{subfigure}
     \hfill
     \begin{subfigure}{0.3\textwidth}
         \centering
         \includegraphics[scale=0.23]{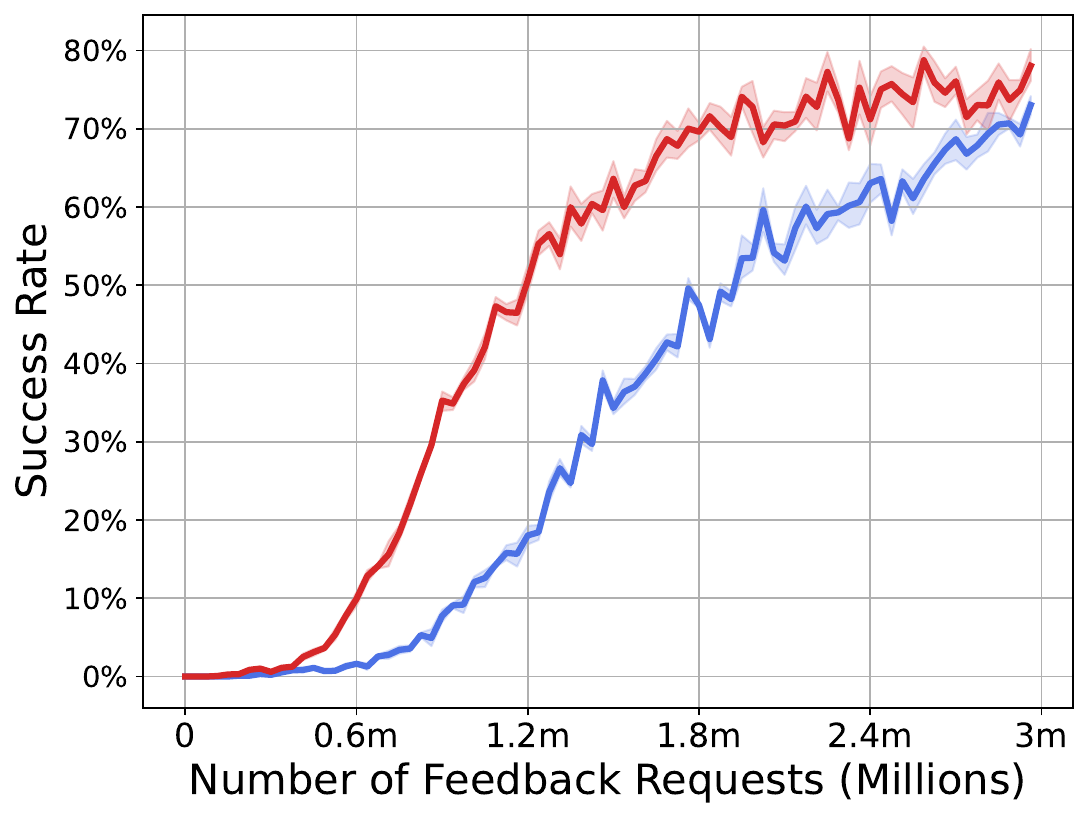}
         \caption{Language-based pragmatic teacher}
         \label{fig:language_comm}
     \end{subfigure}
     \hfill
     \begin{subfigure}{0.3\textwidth}
         \centering
         \includegraphics[width=1.0\textwidth]{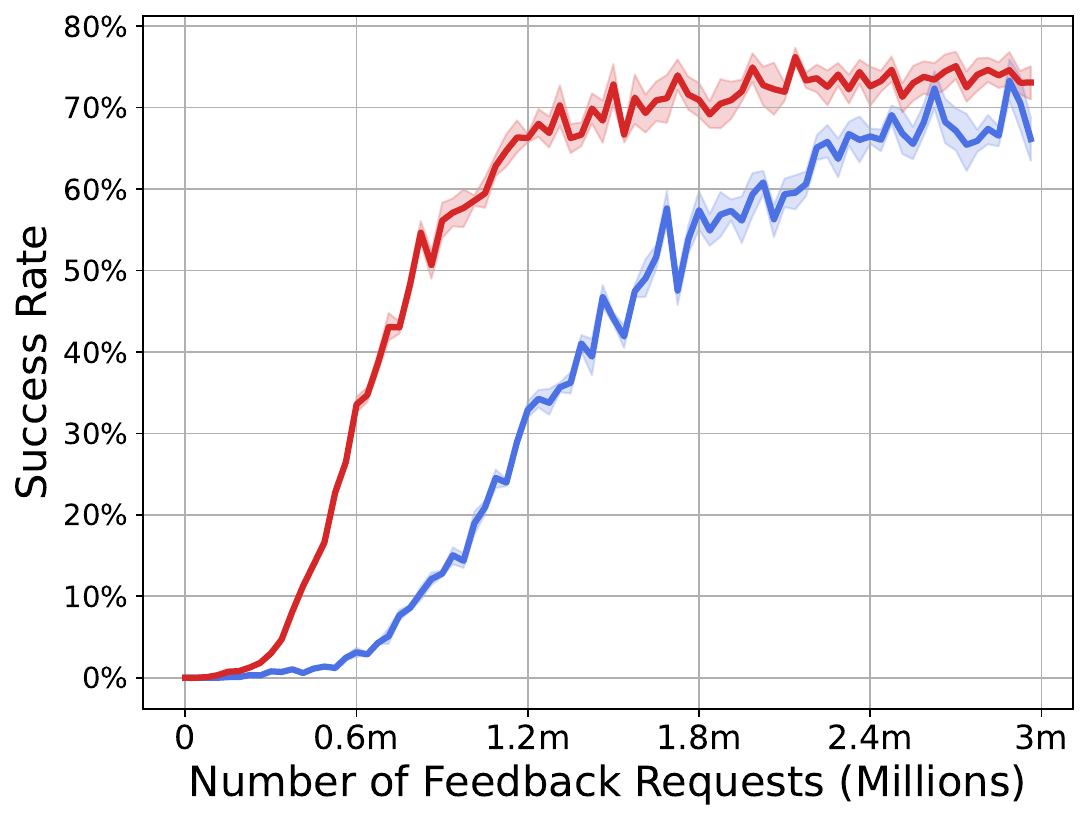}
         \caption{Performance-based pragmatic teacher}
         \label{fig:performance_comm}
     \end{subfigure}
\caption{CEIL with and without minimizing next-episode communication effort.}
\vspace{-0.5em}
\label{fig:ceil_comm}
\end{figure}

\autoref{fig:ceil_comm} shows the results of \ouralgo with and without minimizing the next-episode communication effort ($J_{\textrm{com}}$). 
The objective term does not noticeably impact the performance with the top-down teacher. 
However, with the pragmatic teachers, it becomes crucial for enhancing communication efficiency because it effectively guides the learner towards uttering increasingly more abstract intentions.

%% file: appendix/a5_implementation.tex
\section{Additional implementation details}
\label{sec:impl}

We implement our baselines and \ouralgo based on the \texttt{stable-baselines3} ~\citep{stable-baselines3} codebase. 
We trained all agents with Nvidia 2080Ti GPUs.
It took about one and a half days to train \ouralgo with a 3M feedback-request budget.
For the imitation learning baselines, we used an learning rate of $10^{-4}$, and for \ouralgo, we used $5 \cdot 10^{-5}$. 
% We set the update frequency of the execution classifier $K$ to be 10,000 and $\epsilon$-exploration for execution to be 0.03.
For \ouralgo, we apply a communication cost of 0.01 per each request for instructive feedback when the student's intention is correct, and 0.05 when it is not. 
The cost for giving evaluative feedback is 0.2 per request.

Below is the CNN architecture of the state encoder of the learner's policy:
\begin{lstlisting}[language=Python]
state_encoder = nn.Sequential(
    nn.Conv2d(n_input_channels, 16, kernel_size=1, stride=1, padding=0),
    nn.ReLU(),
    nn.Conv2d(16, 32, kernel_size=3, stride=2, padding=0),
    nn.ReLU(),
    nn.Conv2d(32, 64, kernel_size=3, stride=1, padding=1),
    nn.ReLU(),
    nn.Conv2d(64, 96, kernel_size=3, stride=1, padding=1),
    nn.ReLU(),
    nn.Conv2d(96, 128, kernel_size=3, stride=1, padding=1),
    nn.ReLU(),
    nn.Conv2d(128, 64, kernel_size=1, stride=1, padding=0), 
    nn.ReLU(),
    nn.Flatten(),
    nn.Linear(1024, 256))
)
\end{lstlisting}

%% file: appendix/a6_teachers.tex
\section{Pragmatic teachers}
\label{sec:prag-teacher}

We consider two heuristically pragmatic strategies.
With the \textit{language-based} strategy, the teacher aims to speak at the same level of abstraction as the learner. 
We define the level of abstraction of an intention as the optimal number of actions required to complete the corresponding task. 
Let $T^{\star}_u$ be the level of abstraction of  intention $u$. 
The teacher samples an intention to return according to the probability distribution $P(u) \propto |T^{\star}_u - T^{\star}_{\hat u}|^{-1}$ where $T^{\star}_{\hat u}$ is the level of abstraction of the learner's proposed intention.
With the \textit{performance-based} strategy, the teacher records of the moving success rate $\rho_{u}$ of the learner in executing each intention $u$.
The returned intention is sampled according to $P(u) \propto \rho_{u}$.
Thus, the better the learner is at executing an intention, the more likely the teacher refers to that intention when instructing it.